# Morality, Machines, and the Interpretation Problem: A Value-based, Wittgensteinian Approach to Building Moral Agents

Cosmin Badea[1,2][0000-0002-9808-2475], Gregory Artus[1,3][0000-0003-0445-0079]

[1] Imperial College London, London, SW7 2AZ, UK
[2]cosmin.badea23@gmail.com; g.artus@ic.ac.uk[3]

**Abstract.** We present what we call the Interpretation Problem, whereby any rule in symbolic form is open to infinite interpretation in ways that we might disapprove of and argue that any attempt to build morality into machines is subject to it. We show how the Interpretation Problem in Artificial Intelligence is an illustration of Wittgenstein's general claim that no rule can contain the criteria for its own application, and that the risks created by this problem escalate in proportion to the degree to which to machine is causally connected to the world, in what we call the Law of Interpretative Exposure. Using game theory, we attempt to define the structure of normative spaces and argue that any rule-following within a normative space is guided by values that are external to that space and which cannot themselves be represented as rules. In light of this, we categorise the types of mistakes an artificial moral agent could make into Mistakes of Intention and Instrumental Mistakes, and we propose ways of building morality into machines by getting them to interpret the rules we give in accordance with these external values, through explicit moral reasoning, the "Show, not Tell" paradigm, the adjustment of causal power and structure of the agent, and relational values, with the ultimate aim that the machine develop a virtuous character and that the impact of the Interpretation Problem is minimised.

## 1 Intelligence and the Interpretation Problem

The need to build morality into machines is becoming urgent (Le Roux 2016) and is by now a burgeoning field in Artificial Intelligence (AI) research. The work of thinkers such as Bostrom (2014), Wallach and Allen (2008), Wallach and Asaro (2016), Yudkowsky (2008), and others (Anderson and Anderson 2011) warns us that advances in the power of machines will very quickly create situations where machines will be able to make decisions that not only were previously the sole preserve of hu-man agents, but which have moral significance. Autonomous, or semi-autonomous military robots are the most obvious example of this, but there are far more mundane uses to which machines may in future be put, where ethical problems may arise, be-cause the likely capabilities of future machines will create moral dilemmas that may not have existed for less capable machines, generated not by machines going rogue and deliberately trying to kill us (though this possibility cannot be ignored), but by machines inadvertently harming us while they carry out instructions we have given them. This critical issue is caused by an equally critical problem which we call the Interpretation Problem (IP). Intelligence plus the Interpretation Problem equals trouble.

By intelligence, we of course do not necessarily mean anything as grand as consciousness or Artificial General Intelligence (AGI), but, rather, the ability to be an effective and creative utility (or function) maximiser, i.e., a machine that is 'clever' at finding ways to achieve the goals we set for it (Russell and Norvig 2003). Modern learning machines have shown great promise in this direction within certain restricted domains such as Chess, Go and Poker, with programs such as Alpha Go and Libratus using learning to develop new strategies to achieve the goals of the game of Go and Poker (i.e., winning), strategies that surprised and defeated human champions and professional players decisively (Silver et al. 2016, BBC 2016, Solon 2017). Conse-quently, it is not too far-fetched to think that in future we will be able to build ma-chines that are such good means-ends reasoners, or goal maximisers, that they will be able to think of creative new ways to achieve the ends that we provide for them, which is where IP becomes a pressing issue.

The **Interpretation Problem (IP)** is the general problem that any rule or goal is capable of being interpreted in an infinite, or at least unspecifiable number of ways, and in the field of AI it leads to the possibility that a highly advanced machine may find novel interpretations of the rules that we give it, interpretations which are not incorrect, in that they can be seen as valid interpretations of the rule, but which are inappropriate in that we do not approve of them.

Bostrom's (2014) paperclip maximiser is an illustration of the risks of such machines. An intelligent goal maximiser programmed with the seemingly innocuous goal of maximising the number of paperclips in its collection might work out that the most efficient means to that end is to steal them, or trick or otherwise con people out of them. Such obviously undesirable methods would be easy to predict and address by building prohibitions into the programming. However, it may then find undesirable means we hadn't thought of, or might use too many resources, or even the wrong type of resources in their manufacture (e.g., using the atoms of its makers as raw material), so we'd have to predict and impose yet more limits in its programming, and so on.

If we want the machine to simply make paperclips according to a pre-specified method or procedure, we could break the task into algorithmic steps and program that into the machine (like in Rule-based AI), but even then, and especially when we want the machine to use its learning ability (like in Machine Learning), IP becomes an issue as the machine finds new ways to achieve its goal, and is a function of the very aspect of intelligent machines that makes them useful to us; their ability to adapt, learn and overcome. We argue that this problem is a practical demonstration of one of the key findings of Wittgenstein's later work in *"The Philosophical Investigations",* that what we mean by words cannot be represented by a rule for its use because no rule can contain the criteria for its own application, and that this is a deeper problem with rules in general. Hence, any rule we represent to a machine cannot be specified in an unambiguous way because any representation of a rule is open to multiple interpretations. With the paperclip maximiser and other innovative goal-maximisers, no matter how many restrictions and caveats we predict and build into its programming by representing them as rules, there is always the possibility that it will find novel interpretations of those rules which will clash with our wider goals and values.

This is the basic structure of IP, and it is worth pointing out at this stage that IP is not just a problem for the sort of AGI or super-intelligent machines that Bostrom's example talks of. Any machine that is both linked causally to the world and can also innovate would expose us to IP. Deep Blue, AlphaGo and Libratus all operate within the completely closed artificial domains of a game and so have no causal relation to the world and could not cause us any trouble, but as Olivia Solon points out in an article about Libratus (Solon 2017):

> *The algorithms that power Libratus are not specific to poker, which means the system could have a variety of applications outside of recreational games, from negotiating business deals to setting military or cybersecurity strategy and planning medical treatment – anywhere where humans are required to do strategic reasoning with imperfect information.*

Thus, it seems that these and similar programs will eventually be causally linked to the real world somehow, either online or through some form of embodiment, and may be used to negotiate business deals, manage defence, cyber-security, medical treatment and other important areas of our lives. Bostrom's Paperclip maximiser had almost omnipotent causal capability, but machines with far less causal power will still be dangerous because of their creativity and ability to develop strategies we could not predict or guard against in advance. The more causal power it has in the real world, the greater the risk, so there will be a lawlike relation between the causal capability or connectivity of a machine and our exposure to IP. Hence, for any intelligent utility maximiser, the degree to which we will be exposed to IP will be proportional to the degree to which the machine is causally connected to the world. We call this relation the **Law of Interpretative Exposure**.

## 2 Moral rules and values

If Wittgenstein is correct, then the Interpretation Problem (IP) is not a problem that can be overcome by formulating more precise, or less ambiguous, rules, but is an inherent problem that applies to all rules, not just rules that direct an agent to optimise a goal. Even a simple instruction or imperative, prescriptive rule, or a proscriptive rule requires interpretation and cannot contain the criteria for its own application. It is in principle impossible to specify any rule in such a way that it cannot be misinterpreted and what we mean by a rule can never be unambiguously represented.

An obvious solution to this problem with intelligent machines, then, would seem to be not to try to predict every possible misinterpretation and close it off with a specific rule, but to program into the machine a set of general values or moral rules that would restrict its behaviour to within limits we find morally acceptable. People often cite general laws like Asimov's three laws of robotics here as an example of the sort of thing that might do the job. Such a project, however, is easier said than done. The history of philosophy is littered with attempts to reduce our moral reasoning to either prescriptive or proscriptive rules, and every attempt appears to flounder on the problem that the application of any moral rule seems to be context sensitive, and one can always find interpretations of those rules that we find morally repugnant. Utilitarian or consequentialist accounts framed in terms of prescriptive rules such as 'Do that which promotes the greatest happiness for the greatest number' create all the standard objections to Utilitarianism, such as that of what we mean by 'happiness'. Do we mean material wellbeing, simple physical pleasure, intellectual pleasure, psychological wellbeing, or some other definition of happiness? Deciding which one we mean will always require us to make a value judgement about how the rule is to be interpreted, since the rule itself cannot tell us. And even if we do specify a definition of happiness, it is still possible to find interpretations of that definition which clash with our values, as Yudkowsky's example of the smile maximiser (Yudkowski 2008) am-ply demonstrates – the machine may simply work out that the best way to make more people happy is to make them smile, and that the most efficient way to make everyone smile is by paralysing their faces into a rictus.

Alternatively, the rule could be interpreted as meaning that we should sacrifice the lives or wellbeing of a minority for the sake of that of the majority, which in some cases may seem right, but in others would be abhorrent, such as harvesting the organs of an innocent minority to find a cure for cancer for the majority. In both types of case, it seems that how we mean the rule is guided by values external to the rule itself and is a judgement that can only be made from a stance outside of the rule. This point will become clearer from our later discussion of games and normative spaces.

Proscriptive rules have fared no better than prescriptive rules in moral theory. De-ontological or duty-based approaches, such as Kant's, or Asimov's laws of robotics place prohibitions on us such as 'do not harm other human beings', but this once again raises the question of what we mean by harm. And even if we can come up with a suitable definition, then we are left with the problem of whether we can install in the machine a sufficient understanding of the way the world works for it to be able to recognise how its actions may have wider effects that may cause harm to humans. It would basically have to know (knowing-that or knowing-how) nearly everything about humans and how they live and what can harm them, and even then, it may innovate new forms of action which create new forms of harm that simply didn't exist before. New technologies and new forms of human action are always creating moral dilemmas which didn't exist before, which force us to make judgements about how such rules as 'Do not Harm' apply, and how we interpret or apply the rule in any novel case can only be determined by values external to our rule, values which our rule is in principle incapable of embodying unambiguously.

There is a sense in which IP has been recognized in various ways by moral philosophers for millennia. In Plato's "Republic" (pt1 §1) we see Socrates demolishing the claim of Cephalus that justice consists of little more than following simple rules such as 'tell the truth' and 'always pay one's debts' by responding with the example of a situation where you've borrowed a friend's knife. One day the friend comes hammering at your door hysterically demanding his knife back so that he may kill a man who has upset him. Should you simply tell him the truth (that you still have his knife) and repay your debt so that he may go and kill the man in his unhinged current state? Clearly not. For Plato, moral rules are context sensitive and moral knowledge/wisdom of how and when to apply those rules requires an intellectual grasp of the abstract form of the good, which itself is ineffable and unrepresentable. His pupil Aristotle (1988) agreed that moral knowledge/wisdom was context sensitive and unrepresentable but rejected Plato's notions that it was a form of ineffable intellectual knowledge, preferring instead to treat virtue as a form of practical wisdom that could only be learnt through long experience and practice in pursuit of the flourishing life – a sort of practical know-how neither derivable from nor reducible to representable rules. Nearly every moral theory ever since the Greeks has found the need to ground our moral rules in some further source of values that is external to them: Utilitarian-ism in our innate appetite for pleasure and aversion to pain; Hume and other Naturalist theories, such as evolutionary ethics, in our biologically evolved natural sympathy for our fellow humans; Kant in the principles of logical coherence as expressed by the categorical imperative; Hegel in the social structures of our communities; Moore in an ineffable or brute intuition…the list is endless. Each one of these models, how-ever, is an attempt to find, once and for all, the criteria, or values, by which our moral rules are to be interpreted, and in this sense, one could argue that they are all attempts to overcome IP by describing some further, or final, rule by which moral rules are to be interpreted. But Wittgenstein's work showed us that this task may be futile, since any further rule will itself need to be interpreted, and so on, simply because that is the nature of rules. To see this, it will be instructive to look at the way rules operate in games.

## 3 Normative Systems as Spaces of Possibility

Many games can be seen as a structure of different types of rules that combine to create spaces of possibility and choice in which players must pursue a specified goal or aim, under certain limitations. For example, the aim of the game of soccer is to score goals, while the limitations players are under are that they are not allowed to use their hands, or 'foul' their opponents, move off-side, etc, and the rules define what counts as a goal, what constitutes handball, off-side etc. So, the rules will consist of prescriptive imperatives that determine the aims or goals of the game, plus proscriptive rules laying out the limitations within which the players must pursue those aims, plus several definitional rules - what Searle would call *constitutive rules* (Searle 1969) - which define what is to count as scoring, handball, off-side etc. Most games seem to conform to this general pattern of the pursuit of goals under limitations within a normative space artificially created by the rules, a space of action. The net effect is that *the rules create an arena, or space of possibilities*, in which players are forced to make choices about how best to achieve the specified goal or aim, given the limitations under which they must play. In this sense, instead of directing players or determining their actions, *the purpose of the rules seems to be to create new forms of choice and new dilemmas* that did not exist before. Even in games like Monopoly, where many of the player's actions are determined by imperative conditional rules of the form 'if x, then do y', the structure of the game is such that it creates moments of dilemma and choice at specific points during the playing out of the game. Hence, games often seem mechanisms which use carefully crafted sets of different types of rules to create normative spaces of possibility, designed to extend and test the way we exercise our creative agency, by enabling new types of action.

One of the important things to notice about such normative spaces of possibilities is that they are, for all practical purposes, spaces of infinite possibilities which can accommodate an almost infinite number of possible strategies and tactics; there are an infinite number of ways to score a goal, a touchdown, a run, or to checkmate one's opponent, etc., so one can never exhaustively specify in advance all the possible tactics, moves or strategies that a rule space makes possible. This is because, as Ryle (2000) showed us, principles of strategy form a distinct set of rules which presuppose the rules of the game but are not derivable from those rules. There is no way that from my knowledge of the rules of chess I could logically deduce or otherwise predict that someone would one day invent the Sicilian Defence, or that the tactics of either bodyline bowling or sledging would be the inevitable outcome of the rules of cricket. So, if tactical principles are not logically deducible from the rules of the game, then the number of possible tactics any game may generate is limited only by the imagination, creativity, and ingenuity of the players, and so, with enough ingenuity on the part of the players, *there are, for all practical purposes, an infinite number of ways to interpret the rules* of most games.

From this we can see that games provide us with a perfect illustration of IP in action. There are an infinite number of possible tactics, strategies or interpretations that stay within the rules of the game, just as there are an infinite number of ways an AI can achieve the goals we set it, no matter how many limitations we place on it – new limitations simply force us to make new choices. Furthermore, just as we saw with AI, sometimes players in a game invent a new strategy that we consider undesirable, or do not approve of, despite its still being legally within the rules of the game. Therefore, the rules of games are continually evolving to take account of novel tactics that, for whatever reason, we think inappropriate or undesirable. But what happens when we make such modifications can be very instructive in revealing yet more complexity in the way rule systems work, because it forces us to ask why we find particular tactics undesirable and also what criteria we use as our guide when we alter the rules.

*Chasing the Spirit of the Rules***.** The usual reason we modify the rules of a game is because a new tactic seems to clash somehow with what might be called *the spirit* or purpose or point of the game, so we alter the rules to outlaw it and so maintain the spirit or values that the game is supposed to express. A good example of this occurred in American football in the first half of the twentieth century, when players invented a new strategy called the 'flying wedge', where the whole team linked arms to form a V-shaped wedge in front of the ball carrier and then charged headlong down the pitch as one. There was nothing in the rules that prohibited such a tactic and there was no way that anyone could have deduced or predicted from the then current rules of the game that this tactic would inevitably be utilised by someone. It was solely the product of the creativity and ingenuity of the players, and it was extremely effective at scoring touchdowns if, as was usually the case, the opposing team could not find a way to break up the formation. It soon became clear, however, that this tactic had to be outlawed for several reasons. It made the game very dangerous and was a bit too effective and made the game more of a battle of brute strength than a game of artistry and skill. The tactic seemed to stifle the game and did not encourage creativity and made the game boring both to play and to watch, and the fact that the flying wedge was outlawed shows that the main reason we invented the game in the first place was to encourage the development of just those things stifled, e.g., creativity, flair, artistry, skill etc. Yet nowhere in the rules of American football are artistry, flair, skill, or creativity mentioned. *The game seems designed to promote certain values, yet nowhere in the rules are those values represented. In this sense, the values according to which we believe the rules of the game ought to be interpreted are external to the game itself, so it seems that the normative space created by the rules is designed to express those values, but those values cannot actually be represented by the rules* –such as 'Always interpret the rules of the game artistically'. This rule itself would have to be interpreted by players, and there are an infinite number of ways to do so.

The rules, then, seem designed to promote, encourage or otherwise generate new types of activity through which players can express certain values in the way they act. We invent games where we express intellectual flair, reasoning, or resilience; games where we can express physical prowess, artistry, and skill; games of perseverance, patience, creativity, strategizing and so on. And in this sense, games create what Hannah Arendt would call *'spaces of appearance'* (1998, pt5), spaces created by a system of rules which allow agents to express who they are and what they value and to so define themselves. The games are always designed with certain values in mind, such as those listed above, but they can never guarantee that players will interpret the rules in line with those values because of the nature of normative spaces themselves – they are spaces for the exercise of creative agency, but the very nature of such spaces dictates that players are free to interpret the rules in their own way according to their own values. And when players interpret the rules in ways that clash with the spirit of the game, we simply change the rules.

Furthermore, sometimes the way that players interpret the rules of the game can teach us new values, by, for instance, inventing a strategy that we disapprove of, because it undermines something we took for granted or hadn't realised we valued until it was threatened or lost. It wasn't until the flying wedge was invented that we perhaps realized quite how much we valued other aspects of the game that we had before taken for granted. So, it is not just that the values that normative spaces express cannot be represented in the rules because it is logically impossible for a rule to contain the criteria for its own application, it is also the case that the values expressed by the normative space cannot be represented in the rules because often we don't even know what those values are until they are threatened. In this sense, *normative spaces create spaces of possibility where we can discover values, as well as express existing values*. Often the values that inform a normative space are not formulable beforehand but taken for granted as part of the background of our form of life.

Much of what has been said about restricted domains such as games would seem to apply to normative spaces more generally. For example, in the taxation system legislators are in a continual arms race with taxpayers (or their lawyers and accountants), where the system is designed to express certain values such as 'paying one's dues', yet certain players express their own values as being that of 'minimising one's liabilities'. The normative space created by tax law can never guarantee that all players will play the game according to the spirit or values that underpin the creation of that space because of the nature of normative spaces. So each time someone comes up with a novel strategy that is within the letter of the law, but which clashes with the spirit of the law, regulators must plug the hole with a new rule, or modify an existing rule. And this situation can never be otherwise because of the very nature of normative spaces. Any rule or system of rules always must be interpreted, which in turn presupposes that the agents operating in that space are capable of creative agency. Hence any normative space presupposes both IP plus the creativity of the agents operating within that space. *Consequently, IP is not only in principle impossible to overcome, but is actually what makes normative spaces possible in the first place.* If there were no IP and no creative agents, then we simply could not have normative spaces.

The same would seem to apply in law more generally and in morality, as rules are continually modified according to values external to the rules in response to novel strategies and developments within and across different spheres of our lives. If this is the case, then the quest that has occupied moral philosophers for centuries, that of finding some final rule, criteria or principle which will guarantee that we interpret moral rules in ways that express the values that underpin them, may be futile, and the quest to represent our values as a set of final rules that will guarantee a machine will interpret the rules we give it in line with our wider values may also be impossible.

If this is all correct, then moral rules create normative spaces for the exercise of creative agency where agents express their values in the way they interpret the rules. In this sense, the moral world is not a space where choices can be determined by the rules of the space but is instead an arena for the expression of character. *The virtues one expresses by the way one interprets the rules of the space is who one is.* Consequently, if the moral world is, as this analysis implies and as thinkers such as Arendt have suggested, such a space of appearance that enables agents to define themselves, then any intelligent machine capable of expressing its values within that space by the way it interprets the rules can never be guaranteed to express the values we approve of, so we cannot exhaustively and in advance program a perfectly moral agent. Rather, the way forward for AI research in this topic may be to try to create machines that express a virtuous character in the way they interpret the rules we give them. Yet, as we have seen, character cannot be programmed in as a set of rules but can only be expressed in the machine's interpretation of the rules we give it. Thus, the task of creating virtuous machines seems to be an extremely difficult one. However, in the remainder of this paper we make some suggestions as to how we might begin to tackle this seemingly intractable and daunting task.

## 4 Tackling the Interpretation Problem

**Mistakes of Intention (MIs) and Instrumental Mistakes (IMs).** To begin, we argue, in line with our other work (Badea & Gilpin 2021) that an artificial agent should have as part of its reasoning mechanism two major components, whether made explicit and distinguished from one another or not. The first component will deal with what it ought to do, imperatives and obligations (given or inferred), and the second one will deal with the facts or beliefs it holds, similarly to the inference engine and the knowledge base, respectively, in expert systems (Jackson 1998). We argue that the mistakes of an artificial moral agent can be split into two categories, corresponding to deficiencies in the two parts of the reasoning mechanism described above, *Mistakes of Intention (MIs)* and *Instrumental Mistakes (IMs)*; and we argue that an AI must include an explicit moral program distinct from its practical reasoning program (also argued in Wallach & Allen (2008)).

*A Mistake of Intention (MI)* is one about the imperatives or obligations that the agent has: about *goals* or *limits*. *Mistakes about goals* are occasions on which the agent errs about aims and goals, whether sourced extrinsically, from another agent or the environment, or intrinsically, by its own reasoning. *Mistakes about limits* concern the errors that have to do with actions it should not perform. For example, *mistakes about goals* are all the ones discussed above in the introduction of the paperclip maximised. These stem from the fact that the specified goal had inappropriate qualifications and the same lack could also lead to *mistakes about limits,* such as when asked to "maximise X using only Y kg of matter", the agent using matter of the wrong nature (like cooking the cat in Havens 2015), or by using excessive resources. Another example is that it could infer subgoals, such as nefariously avoiding being turned off, like Hal in "2001: A Space Odyssey". Even if we tried to overcome MIs by programming a moral framework based on a rule-based positive account, whether deontological, consequentialist, or otherwise we would still encounter such mistakes (as we show in Bolton et al. 2022). This is because there might be many circumstances we will not have accounted for (as we show in Post et al. 2022), but most importantly, remember the consequences of IP (above and in our work from 2017): different interpretations are always possible and thus, the spectre of satisfying the literal specification given while not bringing about the intended outcome always looms darkly above us. This kind of behaviour (called *Specification Gaming* by Krakovna et al. 2020), a subspecies of MIs, makes an alternative to these approaches, a way of mitigating IP, glow ever more brightly. We propose this alternative to be *Explicit Moral Reasoning implemented through the "Show, not Tell" (S, not T) method using Values.*

*An Instrumental Mistake (IM)* refers to issues with the part of the agent that deals with facts or beliefs about the world (its knowledge base), similarly to the "*failure of understanding*" in Kantian terms. This can happen if the agent does not correctly understand or predict how the world works, when it is making a factual or empirical error in its reasoning. This type of mistake could occur due to a lack or failure of the agent's common-sense knowledge, and its sources might be false beliefs, incorrect facts, or inappropriate/incomplete understanding of consequences of actions etc.

*Explicit Moral Reasoning for MIs.* As we have argued, the structure of a normative system together with the goals that are usually present in its rules allow us to see parallels between practical reasoning in AI (of which moral reasoning is a part) and human-centred normative systems, for example game-playing or law. Even if we somehow overcame or minimised the IMs (improving its sensors, effectors, common-sense reasoning etc) the agent might still commit MIs, as seen above, and this illustrates why *we must attend not only to the agent's practical understanding of the world, to ward against IMs, but must also build into its reasoning system an explicit moral component, to ward against MIs.* An ontology of the parts required for good decision-making, arguing the same, is attempted elsewhere (Badea & Gilpin 2021).

*Practical and Moral Goals.* Artificial agents can be understood as having two types of goals: *practical goals*, which are almost always explicitly provided through goals it has, and *moral goals*, which are almost never explicit (sometimes not even present), having more to do with the external point, or values of the game, than its rules. As we have seen, practical goals are immediately vulnerable to IP, especially when given in the form 'do X'. We have also seen that any finite limitations, or specific behaviours we place in the same form as part of the rules of the game are not enough to keep the agent on track with being moral, being themselves subject to IP, so we need an explicit approach for moral behaviour. Thus, *we cannot rely on moral behaviour to come as a side effect of purely (non-moral) practical goal-driven behaviour.* The practical, traditional, goals the AI has are part of the specifications of the decision problem, just like they are part of the rules of a game, but as in any game the player, or agent, can come up with ways of acting within the rules of the game but against the external point of the game, because this is not explicitly represented, so we need another mechanism to keep the agent tethered to this external point. To this end, we propose the use of active moral considerations, *values*, to inform the moral reasoning and build a *character*. To implement this, we could employ moral goals built around values. To avoid the pitfalls of purely practical goals, these values should be *explicit* and *efficacious*, that is, be directly present in the agent's reasoning, and have a material impact upon the decision making of an agent in any relevant situations it acts in. We could then have the agent prioritise these moral goals over the practical goals, ensuring that the former are not overruled by the latter.

**Interpretative Exposure (Artus-Badea) Law and Causal Power.** We have seen that there is a lawlike relation (which we over-indulgently call “the Artus-Badea law”) between the amount of causal power a machine has and the degree to which we are exposed to IP. There is a qualitative difference between the consequences of behaviour by disembodied agents with purely digital causal power, and that of embodied fully autonomous agents with very capable sensors and effectors. The more restricted the causal power of the agent, the less unwanted effects it can have on the world, both quantitatively, as the domain it acts in is restricted, and qualitatively, as it can do less impactful actions with minimal (or no) sensors and effectors. Thus, *we propose that we adjust the causal power we build into an agent in the design process to the amount which we believe our reasoning mechanisms can successfully handle* (For example, some argue that a moral advisor is the most causal power to safely start with). Secondly, we can split the process or system of reasoning into multiple parts and focus on one while fixing the others (as advised in Badea & Gilpin 2021). For example, mirroring the types of mistakes an AI can make, MIs can be covered by a moral (or practical) reasoning system and IMs by a purely instrumental reasoning system. We could focus on the moral reasoning system and provide the instrumental understanding required ourselves by hardcoding it into the agent, thus eliminating any IMs, and focus on examining its moral reasoning system, looking for and addressing MIs (as in the moral decision-making framework we describe in Badea 2022). This is useful, as we do not yet have the sophisticated sensors and effectors required for human-level perception, common sense reasoning or understanding of the world and this method would allow us to focus on the moral reasoning without having to deal with all that.

**Evaluation and Building Valuable Character.** For evaluation, we might be tempted to require that the agent be able to justify its solutions to action selection problems using an exhaustive enumeration of actions based on a causal account. To do this, we might want it to represent explicitly the reasoning behind the decisions it makes purely in the form of imperatives like “if S, do X then Y”, but we might not be able to accurately judge it by doing only this if IP holds, because there may be no way for us to have the agent motivate its moral decisions by having it present only a conclusive chain of imperatives or obligations to follow, such as “One is moral in a situation of type S if and only if one first performs action X, followed by action Y etc". This is even before we mention the difficulty of deciding upon and including in advance, or learning at runtime, such a chain of imperatives for any possible variation upon the situation to be encountered. We could then get the agent to give us some (any) reasoned answer to start with, and then evaluate and modify the reasoning system we have, *a posteriori,* either manually or automatically through a learning process, without expecting it to be able to identify in a representable way the solution to complicated ethical problems *a priori* before this testing and training.

Regardless, we could require that it be able to *give a justification* for why it chose to act in a particular way, for us to gauge whether it is indeed moral or not. It could give an explanation based on the decision-making system it contains, the steps it follows in its reasoning (as we saw above) or, perhaps, the relevant values that informed its choice. Maybe it can show us the relevant values or considerations that it holds, and by looking at this and the agent's step-by-step reasoning and implementation, we can piece together how the moral decision making occurred and pass judgement on the whole package and the types of behaviours it exhibits, on this 'character' that it would have. This is another reason why building a moral agent that we can then evaluate could be centred around values, or virtues, aspects of the core of its 'character'.

## 5 On Implementing a Value-based Virtuous AI

From a technical point of view, this 'character' could be made of two parts, an interrelated mechanism of explicit values and considerations that we might call the *moral paradigm*, and a corresponding *moral reasoning engine* that handles their application. An essential question that we are faced with is, then, what moral paradigm should we put in? As we have mentioned before, we cannot straightforwardly use classical versions of utilitarianism or deontology, due to IP and their rule-based nature. So how, then, do we get it to understand that which keeps human rule-following behaviour within acceptable moral bounds?

**Show, not Tell (S, not T) and Values as Vehicles for Meaning.** We propose a way of explaining abstract philosophical concepts (and moral paradigms) to agents, different from either defining what rules they consist of or attempting a direct representation of their content. If conveying meaning through any representational medium is subject to IP, then these conventional methods will be eternally enslaved to IP and thus to fatal ambiguity, a terrible bug in our program. A glimpse of a solution can be seen in the "Philosophical Investigations", itself a work that attempts to take us on a journey to the meaning behind the words, in the absence of any symbolic representation to convey that meaning unambiguously. We thus propose this same paradigm of **"Show, not Tell"** (S not T), but how do we *show* the meaning to the agent? Once again, we take a page from Wittgenstein's book (literally) and get the agent to understand our meaning by taking it on a journey, through a process. We start this process by engineering for it a device to help it indirectly understand what to do and how we want it to be, one which we have extensively advocated for above, a vehicle for transporting meaning between agents: *values*. By **values** we mean high-level concepts that are relevant considerations during decision making. These could be virtues, character traits ("honesty"), or concepts that are of moral importance ("property") or even morally neutral practical considerations. We argue that values are the tether to the external point of the game, crystallising what we want from the behaviour of the agents in the game, or in the moral situation. This is supported by arguments from *Virtue Ethics*, and in particular Aristotle's (1988) connection between virtues and practical wisdom (Phronesis). *Our theory is therefore equally applicable to practical reasoning of any type, not only moral reasoning, and to any kind of agents, not only AI*.

*Leaning into Ambiguity.* An important reason behind the usefulness of values is their ambiguity, and the fact that they are multiply realisable. That is, they can be embodied, or promoted, by different actions (even in the same context) and can be adhered to by a plethora of behaviours. But if we wish to keep values as ambiguous as possible, then, don't we give the agent leeway to perform differently to a specific desired behaviour? Was not getting the agent to do exactly what we wanted the whole point of building our AI in the first place? To this, we would say that our goal in building AI is indeed obtaining some behaviour, but what we want from an (even minimally) independent practical agent is not for it to "do action X, followed by Y, and then stop", but, rather, to achieve a complex goal while acting in a virtuous way. *The purpose of a value-based approach would be to allow the agent freedom to improvise in practical terms, while ensuring that it exhibits a certain character while doing so,* and thus now ambiguity is a feature, not a bug.

**Character as Goal, Using Structure, and Relational Values.** The whole point behind IP is that, in a sufficiently complex environment any goal or behaviour or piece of meaning, when conveyed in a representational way (using natural language, code, programming languages etc), needs to be interpreted and therefore can be misinterpreted and misunderstood. If we recognise this, then we can move forward by attempting to convey the meaning that we desire using more abstract constructs (values), attempting to show indirectly, rather than tell. To start with, some values might be amenable to clear definition for a particular purpose, such as when building domain-specific AI, and they could form anchors (for example 'property' and its definitions in law). The idea here is that there can be some concrete starting point in terms of programming in values, and we can then get it to act in the spirit of these values. This we could do, for example, by *giving the agent the goal of solving the problem of becoming a certain type of character,* by building an explicit value-based moral paradigm into its reasoning, and then examining the results we get and iteratively fine-tuning it (manually or automatically) based on its behaviour. In this way, the machine would be using its creative ability to maximise the primary goal of, say, being trusted by its trainers, or being considered honest and so forth, so we would be getting the machine to do some of the work of mitigating IP for us by *giving it the goal of exhibiting certain virtues*.

*The Structure of Values*. Explicitly having values in the reasoning is beneficial, but to further add accuracy of meaning to our system we can leverage the structure we place the values into, and the relative preferences/interactions in the moral paradigm (Badea 2022), and choosing a structure that furthers our desired use of values seems essential. Even while being aware of IP, we still need a representable way of building the reasoning, as any kind of programming, whether rule-based or learning-based, starts with symbolic representation (pseudo-code, code, specification etc). Due to IP, this representation cannot come (for perfect accuracy) in the form of conventional rules, imperatives, or goals, and so another source of precision is this structure, as the glue that holds values together and the arena in which they can perform. An example of doing this, from game theory/deontic logic/Answer Set Programming, is using a preference relation to classify values based on their relative importance into some ordered structure. We demonstrate a framework for building moral paradigms, with a preference ordering for values, based on a qualitative difference between values arranged in a structure of hierarchical layers, called *MARS*, in Badea 2022.

*Relational Values.* Another way of adding accuracy to the meaning of the solution is by tailoring the type of values we use. We have mentioned that the values used could be virtues, but instead of having a set of simple values, such as 'Honesty', we could use *human-dependent values*, such as 'Being trusted'. We call such values 'relational values' because they are intrinsically relative to another agent. For example, 'being trusted' is a relational value, relative to a particular human because it is measured subjectively through their opinion on the agent's trustworthiness. Thus, it does not need to understand what the values are in isolation, with no quantifiable evaluation of success, but rather in relation to us, measurably. Most attempts to achieve value alignment look at implementing the same rules we seem to follow into the machine (Taylor et al. 2016, Soares 2016). But just as we can misinterpret the rules, so could a machine, and thus we could instead focus on aligning the interpretation of the rules, through these relational values, turning us into moral exemplars for the virtuous machines (as we demonstrate with AI for Medicine in Hindocha & Badea 2022).

We have argued throughout this paper is that there is an important distinction between rules and values. Moral rules, in the form of practical imperatives for example, correspond to the rules of the game, while values, as we have envisaged them, correspond to the spirit of the game, the external point of the game. This external point can only be shown, not said (not accurately transmitted through a representational medium). The tensions that arises between following the rules of the game and the spirit of the game, between the moral rules we give the artificial agent and the kind of behaviour we want it to achieve, stem from the Interpretation Problem. For this reason, and others, we propose the use of values as the core in obtaining moral behaviour. Since it is the spirit of the rules that needs to be acted upon, we propose the process of building artificial moral agent be done through a value-based approach, thus getting our agents to aim at developing a character of which we can approve. Such an anthropocentric, value-based paradigm allows us to train agents that remain tethered to the spirit of our rules and our values, to evaluate our agents as we can one another, and to set ourselves up as moral exemplars for the virtuous machines.